%% file: main.tex
\pdfoutput=1
\documentclass[11pt,a4paper]{article}
\usepackage[hyperref]{emnlp2020}
\usepackage{mathpazo, times}
\usepackage{latexsym}

\usepackage{booktabs}
\usepackage{tabulary}
\usepackage{pgfplots}
\usepackage{multirow}
\usepackage{natbib}
\usepackage{xurl}
\usepackage{breakurl}
\usepackage{hyperref}
\hypersetup{pdftex}
\usepackage{xcolor}
\usepackage{amsmath, amsfonts, mathtools, amssymb}
\usepackage{algorithm}
\usepackage[noend]{algpseudocode}
\pgfplotsset{compat = newest}

\usepackage{microtype}

\aclfinalcopy 
\def\aclpaperid{17} 

\DeclareMathOperator*{\argmin}{arg\,min}
\DeclareMathOperator*{\argmax}{arg\,max}
\def\R{\mathbb{R}}
\def\rr{\boldsymbol r}
\def\ss{\boldsymbol s}
\def\xx{\boldsymbol x}

\title{Sparse Optimization for Unsupervised Extractive Summarization of Long Documents with the Frank-Wolfe Algorithm}

\author{Alicia Y. Tsai \\
  University of California, Berkeley \\
  \texttt{aliciatsai@berkeley.edu} \\ \And
  Laurent El Ghaoui \\
  University of California, Berkeley \\
  \texttt{elghaoui@berkeley.edu} \\}

\date{}

\begin{document}
\maketitle
\begin{abstract}
  We address the problem of unsupervised extractive document summarization, especially for long documents. We model the unsupervised problem as a sparse auto-regression one and approximate the resulting combinatorial problem via a convex, norm-constrained problem. We solve it using a dedicated Frank-Wolfe algorithm. To generate a summary with $k$ sentences, the algorithm only needs to execute $\approx k$ iterations, making it very efficient. We explain how to avoid explicit calculation of the full gradient and how to include sentence embedding information. We evaluate our approach against two other unsupervised methods using both lexical (standard) ROUGE scores, as well as semantic (embedding-based) ones. Our method achieves better results with both datasets and works especially well when combined with embeddings for highly paraphrased summaries.
\end{abstract}

\input{1-introduction}
\input{2-methodologies}
\input{3-experiments}

\input{4-results}
\input{5-conclusion}

\section*{Acknowledgments}
The authors would like to thank Richard Liou, Tanya Roosta, and Gary Cheng for their constructive discussions on drafts of this
paper and SumUp Analytics for providing the dataset.

\bibliographystyle{acl_natbib}
\bibliography{reference}

\input{6-appendix}

\end{document}

%% file: 1-introduction.tex
\section{Introduction}

With the overwhelming increase of digital information, automatic text summarization has become important for many applications such as financial reviews, medical articles, etc. Manually summarizing this amount of information takes a considerable amount of time and effort. This has motivated the study of efficient and reliable automatic text summarization methods. The task of automatic summarization is the process of generating a condensed version of a text that best describes the original one \cite{Hahn:2000, luhn}. The two mainstream approaches in the field of automatic summarization are \textit{extractive} and \textit{abstractive}. 

Extractive approaches generate summaries by selecting a subset of informative words, phrases, or sentences directly from the source text. In contrast, abstractive approaches use linguistic methods to decompose and build a semantic representation of the text and use natural language generation techniques to generate a summary \cite{chopra:2016, Nallapati:2016, Zeng:2016, rush:2015}. In recent years, neural network architectures have made abstractive summarization popular. However, abstractive approaches are generally harder to develop as they require high performing natural language generation techniques, which is also an active research field. Besides these two categories, mixed strategies that combine both extractive and abstractive approaches have also been proposed in recent literature \cite{Peng:2019, cao:2018, see:2017}. Previous work in extractive approaches to summarization include statistical \cite{saggion:2012, Das07asurvey, Goldstein:1999, Kupiec:1995, Paice:1990}, graph-based and optimization-based ones. The graph-based approaches treat text as a network instead of as a simple bag of words and use graph-based ranking methods to generate a summary \cite{Erkan:2011, Ouyang:2009, mihalcea-tarau-2004-textrank}. Optimization-based methods use techniques such as sparse optimization \cite{Yao:2015, elhamifar_sparse_2013}, integer linear programming (ILP) \cite{liu:2013, Berg-Kirkpatrick:2011, woodsend:2011, gillick:2009} and constraint optimization \cite{Durrett:2016, McDonald:2007} to reconstruct the summary.

In this work, we focus on extractive summarization for long documents. Performing automatic text summarization for long documents is especially challenging as obtaining high quality human summaries for long documents is often quite costly and time consuming. Recent works on extractive summarization have been focusing on neural network architectures \cite{nallapati_summarunner_2017, cheng_neural_2016}. Although these methods are successful in generating summaries for short documents, they often have difficulties with long input sequences \cite{shao_generating_2017}. 

Most recent works have started to investigate neural extractive summarization methods for long documents \cite{xiao_extractive_2019, wang_integrating_2017}. However, these methods are supervised and require high quality training data in order to train the neural network models. This creates challenges for domains that do not have massive training datasets. \citet{kedzie_content_2018} compared recent neural extractive summarization models across different domains including news, personal stories, and medical articles. They found that many sophisticated neural extractive summarizers do not have better performance than those consisting of simpler models, and that word embedding averaging performs equally or better than RNNs or CNNs for sentence embedding. This suggests that a simpler model combined with pre-trained word embeddings show promise for summarizing long documents in domains that have few or no training data.

In this work, we propose an unsupervised method for extracting long documents based on a sparse optimization framework and solve it using a dedicated Frank-Wolfe algorithm, which can be combined with pre-trained word embeddings to construct a distributive input representation. Our work is based on the previous work of \citet{cheng_frank_wolfe} but designed specifically for the summarization task. The proposed framework is an unsupervised model that is efficient and does not require a training corpora, as typical supervised solutions would require. We test our method on two datasets that contain long documents, \textsc{2019 Financial Outlooks} and \textsc{Classical Literature}, and compare it against two baselines: sparse subspace clustering (SSC) and TextRank. The experimental results show that our method gives a higher ROUGE score than our baseline for both datasets. In particular, when combined with sentence embedding, our method gives a higher semantic ROUGE score when evaluated on paraphrased summaries. Moreover, we show that our method is computationally more efficient compared to others.



%% file: 2-methodologies.tex
\section{Methodologies}
\label{sec:methodologies}

\paragraph{Notation}
We denote $X_{(i)}$ and $X^{(i)}$ as the $i$-th row and $i$-th column of a matrix $X$ respectively. The matrix $X_t$ denotes the value of $X$ at iteration $t$ while $[X_t]_{(i)}$ and $[X_t]^{(i)}$ denotes the $i$-th row and $i$-th column of $X_t$. The sum $\sum_{i=1}^n \|X_{(i)}\|_2$ is the $L_2$ norm group LASSO constraint. The norm $\|\cdot\|_F$ is the Forbenium norm.

\subsection{Sparse auto-regressive problem}
Extractive summarization aims at finding a minimal set of representative sentences of the original document that effectively summarizes the entire document. Let $A \in \R^{d \times n}$ be the data matrix that represents the document where each column of $A$ represents a sentence in the source document. Here, $d$ is the number of features for each sentence, and $n$ is the number of sentences in the source document. The source document $A$ is written as
\begin{align*}
    A \triangleq
    \begin{bmatrix}
    a_1 & a_2 \cdots & a_n
    \end{bmatrix}
\end{align*}
where the column vector $a_i$ is a sentence in the source document.
Finding the set of representative sentences assumes that the source document $A$ can be approximated by a sparse combination of sentences in the document:
\begin{align*}
    A \approx a_1 x_1^T + a_2 x_2^T + \cdots + a_n x_n^T
\end{align*}
The column vector $x_i$ is a decision variable to be learned. Our goal is to select $k$ sentences whose corresponding decision variable $x_i$ is non-zero. If we write it in a matrix form with $x_i^T$ being the row of the matrix variable $X$, we can formulate the above problem as an auto-regressive problem of the form:
\begin{equation}
\begin{aligned}
\label{eq:def-pb}
    \min_{X} \quad & \|AX - A\|_F^2 \\
    \mbox{s.t.} \quad  &\|v\|_0 \le k  \\ 
    & v_i = \|X_{(i)}\|_2, \;\; \forall i \in [n]  \\
    & X \ge 0 
\end{aligned}
\end{equation}
where $X$ is row-sparse. Here,  $X_{(i)}$ represents the $i$-th row of the matrix variable $X$, $v_i$ is its norm, and the constraint is written in terms of the $L_0$-norm (cardinality, or number of non-zero entries) of $v$, effectively forcing at least $n-k$ entire rows of $X$ to be zero, thereby singling out a short list of at most $k$ sentences that well represent the whole data set. 

The above problem is non-convex and hard to solve but can be well approximated by the so-called $L_1$-norm heuristic, leading to a convex approximation:
\begin{equation}
\begin{aligned}
\label{eq:def-pb-convex}
    \min_{X} \quad &\|AX - A\|_F^2 \\
    \text{s.t.} \quad & \|v\|_1 \le \beta \\
    & v_i = \|X_{(i)}\|_2, \;\; \forall i \in [n]  \\
    & X \ge 0 
\end{aligned}
\end{equation}
where $\beta$ is a hyper-parameter and indirectly controls the row-sparsity (number of non-zero rows). Note that the model simply uses the $L_1$-norm of vector $v$ in \eqref{eq:def-pb} to approximate the cardinality constraint on $v$. If $X^*$ is the solution of problem \eqref{eq:def-pb-convex}, then columns in the data matrix $A^{(j)}$ that correspond to the non-zero rows of $X^*$, $X_{(j)} \neq 0$, are the selected sentences.

\subsection{Frank-Wolfe unsupervised extractive summarization}
The Frank-Wolfe (FW) or conditional gradient algorithm is an iterative first-order optimization algorithm for constrained convex optimization \cite{frank_wolfe_1956}. Although the algorithm was introduced over half a century ago, it has experienced a revival in recent years due to its projection-free iterations and broad applications in machine learning \cite{jaggi2013revisiting}. 

The FW algorithm solves a general constrained optimization problem of the form $\min_{x \in \mathcal{D}} f(x)$,
where the convex function $f$ is differentiable and $L$-Lipschitz and the domain $\mathcal{D}$ is a convex compact set. At each iteration, the FW algorithm requires solving a linear approximation to the objective function over the domain, often referred to as a \textit{linear minimization oracle} (LMO), and then updates the solution accordingly.  At each iteration, we first calculate the gradient, solve the LMO problem to find a descent direction, calculate the step size by line search or by $\frac{2}{t+2}$, and update the estimate.  Algorithm \ref{alg:fw} summarizes the FW process.

\begin{algorithm}
\caption{Frank-Wolfe algorithm}\label{alg:fw}
\begin{algorithmic}[1]
    \State Let $t \gets 0$ and $\xx_0 \in \mathcal{D}$
    \For {$t = 0, 1, \dots, $}
        \State $\ss_t = \argmin\limits_{\ss \in \mathcal{D}} \langle \ss, \nabla f(x_t) \rangle$
        \State Set step size $\rr_t \gets \frac{2}{t+2}$ or
        \State $\rr_t \gets \argmin\limits_{\rr \in [0, 1]} f(\xx_t + \rr (\ss_t - \xx_t))$
        \State Update $\xx_{t+1} = \xx_t + \rr_t (\ss_t - \xx_t)$ \label{alg:update-next-x}
    \EndFor
    \State \Return $\xx_t$
\end{algorithmic}
\end{algorithm}
Unlike other descent methods for constrained optimization that require a projection step at each iteration, the FW algorithm is a projection-free algorithm and only needs to solve the LMO. Applying the FW algorithm to our sparse constrained optimization problem \eqref{eq:def-pb-convex} results in an unsupervised method for extractive summarization. Algorithm \ref{alg:fw} is written in terms of vector variable $x$; however, it is straightforward to extend it to the matrix variable $X$. The algorithm starts with $X_0 \gets 0$, meaning no sentence is selected at first. Then the algorithm greedily selects one sentence at each iteration. The algorithm terminates once $k$ rows of $X_t$ are non-zero (dense) or when the algorithm converges to a row-sparse solution with $k^* < k$ non-zero rows. The complete algorithm of Frank-Wolfe unsupervised extractive summarization is outlined in algorithm \ref{alg:fw-sum}. Next, we explain the details of the algorithm and provide an efficient gradient calculation scheme.

\paragraph{Linear minimization oracle}
The algorithm requires solving the LMO at each iteration. The solution of the LMO $S_t$ specifies the direction of descent at each step.
\begin{align*}
    S_t = \argmin_{S \in \mathcal{D}} \langle S, \nabla f(X_t) \rangle
\end{align*}
Because of the group LASSO constraint in \eqref{eq:def-pb-convex}, the solution matrix $S_t$ is a rank-1 matrix. The non-zero row of $S_t$ is chosen based on the maximum $L_2$ norm of the gradient's rows:
\begin{align*}
    j = \argmax_i \| [\nabla f(X_t)]_{(i)} \|_2 
\end{align*}
We denote the non-zero row of $S_t$ at row $j$ as $[S_t]_{(j)}$. The magnitude of $[S_t]_{(j)}$ is $\beta$ and the direction is chosen to minimize the inner product:
\begin{align*}
   \big[ S_t \big]_{(j)} = - \beta \frac{\big[ \nabla f(X_t) \big]_{(j)}}{ \big\| \big[ \nabla f(X_t) \big]_{(j)} \big\|_2}
\end{align*}
The algorithm produces sparse and low-rank iterates since at most one extra row of $X$ becomes non-zero in each step by the addition of $\rr_t S_t$.

\paragraph{Efficient gradient calculation}
The algorithm requires calculating the gradient at each iteration. The gradient of the objective function in \eqref{eq:def-pb-convex} is:
\begin{align*}
    \nabla f(X) = 2(A^TAX - A^TA) = 2(KX - K)
\end{align*}
The matrix $K = A^TA$ can be calculated once and used throughout the algorithm. Explicitly calculating the gradient is expensive due to the matrix-matrix product $(\text{naively } \mathcal{O}(n^3))$. However, the structure of the problem allows us to efficiently calculate $KX$ at each iteration. From line \ref{alg:update-next-x} in Algorithm \ref{alg:fw}, we know that $X_t$ is a weighted average of $X_{t-1}$ and a rank-1 matrix $S_{t-1}$:
\begin{align*}
    X_t &= (1 - \rr_{t-1}) X_{t-1} + \rr_{t-1} S_{t-1}
\end{align*}
This suggests that $(KX)_t$ is a weighted average of $(KX)_{t-1}$ and $KS_{t-1} = K^{(j)} [S_{t-1}]_{(j)}$. $K^{(j)}$ is the $j$-th column corresponding to the $j$-th non-zero row of $S_{t-1}$. Since $(KX)_{t-1}$ is known at iteration $t$, we are only required to calculate $K^{(j)} [S_{t-1}]_{(j)}$, which is extremely fast $(\text{in } \mathcal{O}(n))$.

\paragraph{Stopping criteria}
The algorithm terminates once $k$ rows of $X_t$ are non-zero (dense) or when $X_t$ converges to a row-sparse solution such that $- \big\langle \nabla f(X_t), S_t - X_t \big\rangle < \epsilon$. Once the algorithm terminates, it returns the $k$ sentences that correspond to the non-zero rows of $X_t$ by \textbf{GetSummary}$(X_t, k)$.

\begin{algorithm*}[!h]
\caption{Frank-Wolfe unsupervised extractive summarization} \label{alg:fw-sum}
\begin{algorithmic}[1]
    \State {\bfseries input} $\beta, k, \epsilon$
    \State {\bfseries initialize} $X_0, (KX)_0, \rr_0, t \gets 0, 0, 0, 1$
    \State {\bfseries compute} $K = A^TA$ or $\Phi(A)$
    \For {$t = 1, 2, \dots$}
        \State $(KX)_t = (1 - \rr_{t-1}) (KX)_{t-1} + \rr_{t-1} K^{(j)} [S_{t-1}]_{(j)}$
        \State $\nabla f(X_t) = 2 \big((KX)_t - K \big)$
        \State $j = \argmax_{j} \big\| \big[ \nabla f(X_t) \big]_{(j)} \big\|_2$
        \State $S_t = 0$
        \State $\big[S_t\big]_{(j)} = - \beta \frac{\big[ \nabla f(X_t) \big]_{(j)}}{ \big\| \big[ \nabla f(X_t) \big]_{(j)} \big\|_2}$
        \State $\rr_t = \frac{2}{t+2}$ or $\argmin\limits_{\rr \in [0, 1]} f(X_t + \rr (S_t - X_t))$
        \State $X_{t+1} = X_t + \rr_t (S_t - X_t)$
        \If {\textbf{NumSent}$(X_{t+1}) = k$ or $- \big\langle \nabla f(X_t), S_t - X_t \big\rangle < \epsilon$}
            \State \textbf{break} \Comment{$k$ rows are non-zero or $X_t$ converges}
        \EndIf
        \State $t = t+1$
    \EndFor
    \State \Return \textbf{GetSummary($X_{t+1}, k$)}
\end{algorithmic}
\end{algorithm*}

\paragraph{Sentence similarity measure}
We note that the gradient of \eqref{eq:def-pb-convex} depends only on the kernel (or, ``Gram'') matrix $K = A^TA$ and not on $A$. This matrix is akin to a similarity matrix, with $K_{ij}$ measuring the similarity between sentences $i$ and $j$. If the matrix $A$ is normalized, $K_{ij}$'s are cosine similarities. As a result, we may replace the matrix $K$ with any matrix $\Phi(A)$ that offers a good similarity measure between sentences. This allow us to incorporate various sentence scoring functions $\Phi(\cdot)$. In this paper, we experimented with two such similarity measures: 1) TF-IDF-like, and 2) sentence embedding.

For TF-IDF-like similarity measure, we use Okapi BM25 \cite{Robertson:2009} to construct the kernel matrix $K$. BM25 and its variants represent the state-of-the-art TF-IDF-like sentence scoring functions. Similarly, any sentence embedding technique can be used to embed the document matrix $A$ in a much lower dimensional space; that is, we can set $K_{ij} = \phi(a_i)^T\phi(a_j)$, with $\phi(a)$ the (low-dimensional) vector representing the sentence $a$. In this work, we use a simple yet effective sentence embedding method called smooth inverse frequency (SIF) \cite{arora:2017} to measure the sentence similarities. In \citet{arora:2017}, the authors show that SIF, a simple weighted average of word vectors modified by SVD, outperforms complex methods such as RNNs and LSTMs. More sophisticated sentence embedding techniques such as neural architectures can also be used here; however, once should also consider the cost of computing the kernel matrix $K$ with such a technique.
 
In the following, the acronyms FWSum-BM25 and FWSum-SIF are used to refer to the corresponding Frank-Wolfe unsupervised extractive summarization method used in conjunction with the BM25 and SIF similarity kernels.

%% file: 3-experiments.tex
\section{Experiments}
\subsection{Datasets}
\citet{dernoncourt_repository_2018} surveyed the current large-scale dataset for summarization. Most of them are relatively short; usually less than 2 pages. To experiment on long documents, we used the recently open-sourced \textsc{2019 Financial Outlooks} and \textsc{Classical Literature} dataset\footnote{\url{https://github.com/SumUpAnalytics/goldsum}}, which contain much longer documents than those surveyed in \citet{dernoncourt_repository_2018}.


\paragraph{\textsc{2019 Financial Outlooks}}
This corpus contains 10 publicly available reports on finance from a number of large financial institutions. Each report ranges from 10 to 144 pages, with a median length of 33 pages. There are no Gold summaries \textit{per se} since the data is not annotated by a human. Hence, we chose to define the gold summaries as the collection of sentences or parts of sentences that appear in bold in the content, or any sentences that are highlighted as an insert within the content. This is a reasonable heuristic as these parts are generally prepared by the authors to highlight the takeaway of the content. 

\paragraph{\textsc{Classical Literature}}
The corpus contains summaries of books that have been summarized by human writers. The corpus contains 11 English-language classical books ranging from 53 to 1139 pages, with a median length of 198 pages. The Gold summaries for each chapter of the book are retrieved from WikiSummary \footnote{\url{http://wikisum.com/w/Main_Page}}. 

\subsection{Baselines}
We compared our method with two other unsupervised extractive approaches; one uses a sparse optimization-based method and the other uses a graph-based method.


\paragraph{Sparse subspace clustering (SSC)}
Sparse subspace clustering (SSC) solves a sparse optimization program on the auto-regressive problem similar to \eqref{eq:def-pb}, called the \emph{self-expressiveness property} of the data \cite{elhamifar_sparse_2013}. This property assumes that each data point can be efficiently reconstructed by a combination of other points in the data and that there exists a sparse representation of the data point. The authors consider a convex relaxation as we did in \eqref{eq:def-pb-convex} since solving the original sparse optimization is in general NP-hard. SSC uses the Alternating Direction Method of Multipliers (ADMM) for solving the sparse optimization problem. In our work, we employ the Frank-Wolfe algorithm on the problem, which is more efficient compared to SSC. Subsequent work tried to speed up SSC \cite{you_scalable_2016} but with an expense of removing the group LASSO constraint that is crucial for our summarization problem. In our work, we are able to preserve the group LASSO constraint and obtain a faster run-time. In our experiment, we used the implementation of \citet{elhamifar_sparse_2013}, which can be found on their website\footnote{\url{http://www.ccs.neu.edu/home/eelhami/codes.htm}}.

\paragraph{TextRank}
TextRank \cite{mihalcea-tarau-2004-textrank} is a commonly used graph-based unsupervised extractive summarization method. It is also very efficient when extracting summaries from a long document. TextRank employs the similar idea of PageRank where vertices in the graph are sentences in the document and edges between two sentences are measured as a function of their content overlap.
    
\subsection{Lexical and semantic ROUGE scores}
We evaluate the systems using the ROUGE-1, ROUGE-2, and ROUGE-L \cite{lin:2004} so as to account for different summary lengths. The raw ROUGE score only measures the lexical overlaps between the generated summaries and the reference summaries. We refer to the raw ROUGE score defined in \citet{lin:2004} as the \textit{lexical ROUGE} and used the implementation of the Python \texttt{rouge} library\footnote{\url{https://pypi.org/project/rouge/}}. When summarizing a long document, humans tend to paraphrase the source document in order to condense and synthesize the information. However, the lexical ROUGE scores are unable to measure the quality of paraphrasing. To address this shortcoming of lexical ROUGE when the summaries are paraphrased, word embedding ROUGE scores \cite{ng:2015} are also used to evaluate the quality of the generated summaries. The word embedding ROUGE scores are more capable of measuring semantic similarity of the words instead of only lexical overlaps. \citet{ng:2015} showed that the embedding ROUGE achieved better correlations with human assessments compared to lexical ROUGE when measured with the Spearman and Kendall rank coefficients on the TAC AESOP summarization dataset. We refer to the word embedding ROUGE scores as the \textit{semantic ROUGE} in our evaluation.

%% file: 4-results.tex
\section{Results and Analysis}
\input{4-table}

In our experiment, we set the number of selected sentences $k$ to be the same as the length of reference summary for all methods. The performance of all methods on \textsc{Financial Outlook} and \textsc{Classical Literature} are shown in Table \ref{tab:results}. As shown in the table, FWSum-BM25 has a similar performance with TextRank although slightly better. This may be explained by the sentence scoring functions used by TextRank and FWSum-BM25. TextRank uses lexical overlaps between two sentences while FWSum-BM25 uses the TF-IDF-like scoring function, which are similar in nature. 

FWSum-BW25 performs especially well when evaluated with lexical ROUGE, highlighting its capabilities of capturing lexical information (measured by unigram and bigram). When evaluated on the \textsc{Financial Outlook} data, FWSum-BW25 and TextRank generally outperform FWSum-SIF, with FWSum-BM25 being the best performing method. Presumably, this is due to the fact that the Gold summaries of the \textsc{Financial Outlook} data are taken directly from the source document without much paraphrasing, favoring sentence scoring functions that directly measure the content overlaps. 

However, when evaluated by the semantic ROUGE on the \textsc{Classical Literature} data, FWSum-SIF start to show promises. The Gold summaries of the \textsc{Classical Literature} data are written by human writers and are highly paraphrased and condensed. As a result, semantic ROUGE is a better measurement for this dataset. As shown in the table, FWSum-SIF starts to outperform other methods by a significant amount. The improvement over the other methods suggests that using embedding in the sentence scoring function allows for comparisons based on the semantics of words sequences.

This results show that different sentence scoring functions may be used based on the nature of the summary. For summaries that are mostly taken from the source document without much paraphrasing, a lexical overlap or TF-IDF-like kernel matrix may be used. For summaries that are highly paraphrased, an embedding-like kernel matrix may be more suitable. Our method is able to work with both.


\paragraph{Computational complexity}
Our method requires an up-front cost of calculating the kernel matrix $K$. Each subsequent iteration requires mostly the LMO and gradient calculation as detailed in section \ref{sec:methodologies}. By exploiting the structure of the problem, we are able to avoid explicitly calculating the full gradient. Furthermore, due to the greedy nature of the algorithm, it terminates when $k$ sentences are selected or the solution converges with $k^* < k$ sentences. This means that the algorithm only needs to execute $\approx k$ iterations; each iteration has a cost linear in problem size. Figure \ref{fig:runtime} compares the algorithm run-time of our method (FWSum-BM25), TextRank and SSC. As shown in the figure, our method is the most efficient among the three, showing its potential for summarizing long documents.

\input{4-plot}

%% file: 4-table.tex
\begin{table*}[!h]
\centering
\begin{tabulary}{\textwidth}{LCCCCC}
    \toprule
    \multicolumn{2}{c}{Lexical ROUGE} & SSC & TextRank & FWSum-BM25 & FWSum-SIF \\
    \midrule
    \multirow{3}{*}{\textsc{Financial Outlook}} & ROUGE-L F1 & 11.88 & 14.43 & 13.92 & \textbf{14.99} \\
    & ROUGE-2 F1 & 2.15 & 3.76 & \textbf{5.17} & 3.05 \\
    & ROUGE-1 F1 & 14.6 & \textcolor{gray}{\textbf{19.94}} & \textcolor{gray}{\textbf{19.9}} & 18 \\
    \midrule 
    \multirow{3}{*}{\textsc{Classical Literature}} & ROUGE-L F1 & 7.48 & 16.27 & \textbf{18.7 }& 13.18 \\
    & ROUGE-2 F1 & 0.38 & 2.54 & \textbf{3.23} & 1.25 \\
    & ROUGE-1 F1 & 9.97 & 19.61 & \textbf{20.2} & 16.58 \\
    \bottomrule
    \toprule
    \multicolumn{2}{c}{Semantic ROUGE} & SSC & TextRank & FWSum-BM25 & FWSum-SIF \\
    \midrule
    \multirow{3}{*}{\textsc{Financial Outlook}} & ROUGE-L F1 & 30.01 & 26.2 & 22.97 & \textbf{34.56} \\
    & ROUGE-2 F1 & 55.56 & 61.43 & \textbf{61.82} & 58.92 \\
    & ROUGE-1 F1 & 43.4 & 48.28 & \textbf{49.97} & 47.73 \\
    \midrule 
    \multirow{3}{*}{\textsc{Classical Literature}} & ROUGE-L F1 &  31.92 & 39.15 & 39.1 & \textbf{46.6} \\
    & ROUGE-2 F1 & 47.73 & 53.35 & 54.1 & \textbf{60.53} \\
    & ROUGE-1 F1 & 38.72 & 44.15 & 42.43 & \textbf{48.18} \\
    \bottomrule
\end{tabulary}
\caption{Lexical and semantic ROUGE performance for \textsc{Financial Outlook} and \textsc{Classical Literature} data. Results that are statistically better are bold faced and results that are statistically indistinguishable are colored as gray. An additional experimental results can be found in appendix \ref{appendix:additional-results}.}
\label{tab:results}
\end{table*}

%% file: 4-plot.tex
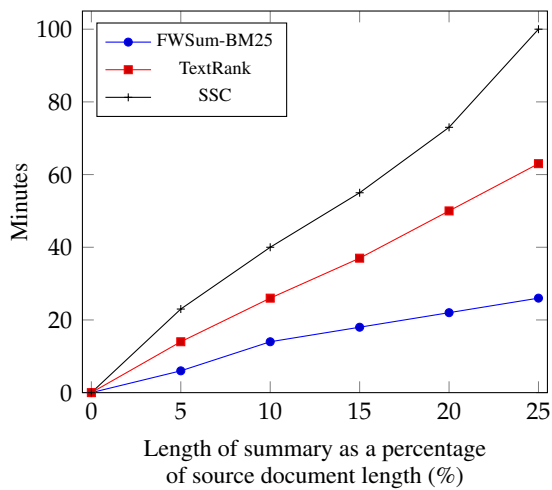
\begin{figure}[!h]
\centering
\begin{tikzpicture}
\begin{axis}[
    width=\columnwidth,
    font=\footnotesize,
    legend style={font=\scriptsize},
    mark size=1.5pt,
    xlabel style={align=center},
    xlabel={Length of summary as a percentage \\ of source document length (\%)},
    ylabel style={yshift=-1.5mm},
    ylabel={Minutes},
    xmin=-0.5, xmax=25.5,
    ymin=0, ymax=105,
    xtick={0,5,10,15,20,25},
    ytick={0,20,40,60,80, 100},
    legend pos=north west,
    grid style=dashed,
]
\addplot coordinates {
( 0, 0 )
( 5, 6 )
( 10, 14 )
( 15, 18 )
( 20, 22 )
( 25, 26 )
};
\addlegendentry{FWSum-BM25} ;

\addplot coordinates {
( 0, 0 )
( 5, 14 )
( 10, 26 )
( 15, 37 )
( 20, 50 )
( 25, 63 )
};
\addlegendentry{TextRank} ;

\addplot[mark=+] coordinates {
( 0, 0 )
( 5, 23 )
( 10, 40 )
( 15, 55 )
( 20, 73 )
( 25, 100 )
};
\addlegendentry{SSC} ;

\end{axis}
\end{tikzpicture}

\caption{Algorithm run-time for FWSum-BM25, TextRank and SSC on the \textsc{Financial Outlook} data. The $x$-axis shows the length of the generated summary (\emph{i.e.} $k$) as a percentage of the source document length (number of sentences in the source document).}
\label{fig:runtime}
\end{figure}

%% file: 5-conclusion.tex
\section{Conclusion}
Unsupervised document summarization has been a challenging task, especially on long documents. In this work, we propose an efficient unsupervised extractive summarization model that is suitable for long documents by employing a dedicated Frank-Wolfe algorithm. Our method allows one to incorporate sentence embedding or any sentence scoring functions that is best suited for the dataset or the application. We evaluate our method and compare it with two other unsupervised extractive summarization methods on two datasets that are much longer than other summarization corpora used in the past. We evaluate the methods on both lexical and semantic ROUGE in order to overcome the shortcoming of lexical ROUGE and to provide a better assessment of the quality of the summaries. We observed that our methods (both FWSum-BM25 and FWSum-SIF) achieve the best results for both datasets and that FWSum-SIF works especially well with summaries that are paraphrased. Our results also motivate the exploration of different kernel functions or embedding methods, which is left as a future work.

%% file: 6-appendix.tex
\appendix
\onecolumn

\section{Additional Experimental Results}
\label{appendix:additional-results}

\begin{table}[!h]
\centering
\begin{tabulary}{\textwidth}{LCCCC}
    \toprule
    \multicolumn{5}{c}{\textsc{2019 Financial Outlook}} \\
    \midrule
    Lexical ROUGE & SMRS & TextRank & FWSR-BM25 & FWSR-SIF \\ 
    \midrule
    R-L F1 & [11.77, 11.99] & [14.28, 14.57] & [13.79, 14.05] & [\textbf{14.87, 15.11}] \\
    R-L Precision & [11.06, 11.28] & [13.05, 13.31] & [12.69, 12.92] & [\textbf{14.63, 14.85}] \\
    R-L Recall & [18.69, 19.04] & [29.06, 29.51] & [\textbf{33.58, 34.09}] & [22.50, 22.97] \\
    R-2 F1 & [02.07, 02.23] & [03.65, 03.87] & [\textbf{05.04, 05.29}] & [02.96, 03.13] \\
    R-2 Precision & [01.63, 01.75] & [02.62, 02.78] & [\textbf{03.44, 03.62}] & [02.37, 02.51] \\
    R-2 Recall & [03.01, 03.23] & [06.65, 07.02] & [\textbf{10.38, 10.92}] & [04.88, 05.25] \\
    R-1 F1 & [14.47, 14.73] & [\textcolor{gray}{\textbf{19.78, 20.10}}] & [\textcolor{gray}{\textbf{19.75, 20.04}}] & [17.87, 18.13] \\
    R-1 Precision & [11.91, 12.15] & [14.77, 15.05] & [13.95, 14.19] & [\textbf{15.56, 15.79}] \\
    R-1 Recall & [20.01, 20.38] & [32.36, 32.81] & [\textbf{36.63, 37.10}] & [24.02, 24.50] \\
    \bottomrule
    \toprule
    \multicolumn{5}{c}{\textsc{Classical Literature}} \\
    \midrule
    Lexical ROUGE & SMRS & TextRank & FWSR-BM25 & FWSR-SIF \\ 
    \midrule 
    R-L F1 & [07.39, 07.56] & [16.15, 16.38] & [\textbf{18.59, 18.81}] & [13.10, 13.26] \\
    R-L Precision & [16.96, 17.23] & [14.03, 14.22] & [15.82, 16.00] & [\textbf{18.04, 18.27}] \\
    R-L Recall & [06.85, 07.00] & [19.02, 19.31] & [\textbf{22.53, 22.81}] & [12.95, 13.15] \\
    R-2 F1 & [00.37, 00.39] & [02.49, 02.58] & [\textbf{03.16, 03.29}] & [01.22, 01.27] \\
    R-2 Precision & [00.68, 00.73] & [01.97, 02.04] & [\textbf{02.87, 02.98}] & [01.64, 01.71] \\
    R-2 Recall & [00.27, 00.29] & [\textcolor{gray}{\textbf{03.39, 03.51}}] & [\textcolor{gray}{\textbf{03.51, 03.66}}] & [01.10, 01.15] \\
    R-1 F1 & [09.87, 10.07] & [19.49, 19.72] & [\textbf{20.10, 20.30}] & [16.48, 16.68] \\
    R-1 Precision & [17.82, 18.10] & [18.43, 18.65] & [16.69, 16.88] & [\textbf{20.33, 20.58}] \\
    R-1 Recall & [07.26, 07.42] & [22.10, 22.44] & [\textbf{26.67, 26.97}] & [14.83, 15.08] \\
    \bottomrule
\end{tabulary}
\caption{Lexical ROUGE performance for \textsc{2019 Financial Outlook} and \textsc{Classical Literature} dataset. We reported the 95\% confidence interval for the results. Results that are statistically better are bold faced and results that are statistically similar are colored as gray.}
\label{tab:lexical-rouge-full}
\end{table}


\begin{table}[!h]
\centering
\begin{tabulary}{\textwidth}{LCCCC}
    \toprule
    \multicolumn{5}{c}{\textsc{2019 Financial Outlook}} \\
    \midrule
    Semantic ROUGE & SMRS & TextRank & FWSR-BM25 & FWSR-SIF \\ 
    \midrule
    R-L F1 & [29.84, 30.18] & [26.02, 26.37] & [22.80, 23.13] & [\textbf{34.36, 34.75}] \\
    R-L Precision & [32.70, 34.01] & [23.78, 24.09] & [20.97, 21.25] & [\textbf{37.83, 38.97}] \\
    R-L Recall & [46.01, 46.33] & [52.45, 52.76] & [\textbf{55.00, 55.33}] & [48.66, 48.95] \\
    R-2 F1 & [55.45, 55.67] & [61.33, 61.53] & [\textbf{61.72, 61.91}] & [58.84, 59.00] \\
    R-2 Precision & [52.59, 52.80] & [\textbf{57.14, 57.33}] & [56.27, 56.46] & [56.17, 56.34] \\
    R-2 Recall & [59.16, 59.45] & [66.46, 66.74] & [\textbf{68.73, 69.00}] & [62.22, 62.49] \\
    R-1 F1 & [43.27, 43.52] & [48.16, 48.40] & [\textbf{49.85, 50.09}] & [45.54, 45.73] \\
    R-1 Precision & [41.31, 41.55] & [44.80, 45.03] & [\textbf{45.92, 46.14}] & [43.23, 43.41] \\
    R-1 Recall & [46.02, 46.33] & [52.45, 52.76] & [\textbf{55.00, 55.33}] & [48.66, 48.95] \\
    \bottomrule
    \toprule
    \multicolumn{5}{c}{\textsc{Classical Literature}} \\
    \midrule
    Semantic ROUGE & SMRS & TextRank & FWSR-BM25 & FWSR-SIF \\ 
    \midrule
    R-L F1 & [31.83, 32.04] & [38.97, 39.33] & [38.89, 39.30] & [\textbf{46.29, 46.91}] \\
    R-L Precision & [21.38, 21.94] & [37.69, 38.16] & [45.74, 46.49] & \textbf{[48.37, 49.44]} \\
    R-L Recall & [33.25, 33.50] & [40.35, 40.58] & [42.19, 42.42] & [\textbf{44.38, 44.62}] \\
    R-2 F1 & [47.62, 47.84] & [52.87, 53.83] & [54.00, 54.20] & [\textbf{60.41, 60.65}] \\
    R-2 Precision & [57.92, 58.19] & [56.11, 56.32] & [55.50, 55.71] & [\textbf{66.39, 66.62}] \\
    R-2 Recall & [40.78, 41.01] & [49.98, 50.19] & [52.84, 53.07] & [\textbf{55.42, 55.66}] \\
    R-1 F1 & [38.62, 38.82] & [44.05, 44.24] & [42.33, 42.52] & [\textbf{48.06, 48.29}] \\
    R-1 Precision & [50.06, 50.38] & [43.99, 44.17] & [42.72, 42.92] & [\textbf{59.80, 60.04}] \\
    R-1 Recall & [31.83, 32.04] & [\textbf{44.38, 44.62}] & [42.19, 42.42] & [40.35, 40.58] \\
    \bottomrule
\end{tabulary}
\caption{Sematic ROUGE performance for \textsc{2019 Financial Outlook} and \textsc{Classical Literature} dataset. We reported the 95\% confidence interval for the results. Results that are statistically better are bold faced.}
\label{tab:semantic-rouge-full}
\end{table}

\twocolumn

%% file: main.bbl
\begin{thebibliography}{40}
\expandafter\ifx\csname natexlab\endcsname\relax\def\natexlab#1{#1}\fi

\bibitem[{Arora et~al.(2017)Arora, Liang, and Ma}]{arora:2017}
Sanjeev Arora, Yingyu Liang, and Tengyu Ma. 2017.
\newblock A simple but tough-to-beat baseline for sentence embeddings.

\bibitem[{Berg-Kirkpatrick et~al.(2011)Berg-Kirkpatrick, Gillick, and
  Klein}]{Berg-Kirkpatrick:2011}
Taylor Berg-Kirkpatrick, Dan Gillick, and Dan Klein. 2011.
\newblock \href {http://dl.acm.org/citation.cfm?id=2002472.2002534} {Jointly
  learning to extract and compress}.
\newblock In \emph{Proceedings of the 49th Annual Meeting of the Association
  for Computational Linguistics: Human Language Technologies - Volume 1}, HLT
  '11, pages 481--490, Stroudsburg, PA, USA. Association for Computational
  Linguistics.

\bibitem[{Cao et~al.(2018)Cao, Li, Li, and Wei}]{cao:2018}
Ziqiang Cao, Wenjie Li, Sujian Li, and Furu Wei. 2018.
\newblock \href {https://www.aclweb.org/anthology/P18-1015} {Retrieve, rerank
  and rewrite: Soft template based neural summarization}.
\newblock In \emph{Proceedings of the 56th Annual Meeting of the Association
  for Computational Linguistics (Volume 1: Long Papers)}, pages 152--161,
  Melbourne, Australia. Association for Computational Linguistics.

\bibitem[{Cheng et~al.(2018)Cheng, Askari, Ghaoui, and
  Ramchandran}]{cheng_frank_wolfe}
Gary Cheng, Armin Askari, Laurent~El Ghaoui, and Kannan Ramchandran. 2018.
\newblock \href {http://arxiv.org/abs/1811.02702} {Frank-wolfe algorithm for
  exemplar selection}.
\newblock \emph{CoRR}, abs/1811.02702.

\bibitem[{Cheng and Lapata(2016)}]{cheng_neural_2016}
Jianpeng Cheng and Mirella Lapata. 2016.
\newblock \href {https://doi.org/10.18653/v1/P16-1046} {Neural {Summarization}
  by {Extracting} {Sentences} and {Words}}.
\newblock In \emph{Proceedings of the 54th {Annual} {Meeting} of the
  {Association} for {Computational} {Linguistics} ({Volume} 1: {Long}
  {Papers})}, pages 484--494, Berlin, Germany. Association for Computational
  Linguistics.

\bibitem[{Chopra et~al.(2016)Chopra, Auli, and Rush}]{chopra:2016}
Sumit Chopra, Michael Auli, and Alexander~M. Rush. 2016.
\newblock \href {https://doi.org/10.18653/v1/N16-1012} {Abstractive sentence
  summarization with attentive recurrent neural networks}.
\newblock In \emph{Proceedings of the 2016 Conference of the North {A}merican
  Chapter of the Association for Computational Linguistics: Human Language
  Technologies}, pages 93--98, San Diego, California. Association for
  Computational Linguistics.

\bibitem[{Das and Martins(2007)}]{Das07asurvey}
Dipanjan Das and André F.~T. Martins. 2007.
\newblock A survey on automatic text summarization.

\bibitem[{Dernoncourt et~al.(2018)Dernoncourt, Ghassemi, and
  Chang}]{dernoncourt_repository_2018}
Franck Dernoncourt, Mohammad Ghassemi, and Walter Chang. 2018.
\newblock \href {https://www.aclweb.org/anthology/L18-1509} {A {Repository} of
  {Corpora} for {Summarization}}.
\newblock In \emph{Proceedings of the {Eleventh} {International} {Conference}
  on {Language} {Resources} and {Evaluation} ({LREC} 2018)}, Miyazaki, Japan.
  European Language Resources Association (ELRA).

\bibitem[{Durrett et~al.(2016)Durrett, Berg{-}Kirkpatrick, and
  Klein}]{Durrett:2016}
Greg Durrett, Taylor Berg{-}Kirkpatrick, and Dan Klein. 2016.
\newblock \href {http://arxiv.org/abs/1603.08887} {Learning-based
  single-document summarization with compression and anaphoricity constraints}.
\newblock \emph{CoRR}, abs/1603.08887.

\bibitem[{Elhamifar and Vidal(2013)}]{elhamifar_sparse_2013}
Ehsan Elhamifar and Rene Vidal. 2013.
\newblock \href {http://arxiv.org/abs/1203.1005} {Sparse {Subspace}
  {Clustering}: {Algorithm}, {Theory}, and {Applications}}.
\newblock \emph{arXiv:1203.1005 [cs, math, stat]}.
\newblock ArXiv: 1203.1005.

\bibitem[{Erkan and Radev(2011)}]{Erkan:2011}
G{\"{u}}nes Erkan and Dragomir~R. Radev. 2011.
\newblock \href {http://arxiv.org/abs/1109.2128} {Lexrank: Graph-based lexical
  centrality as salience in text summarization}.
\newblock \emph{CoRR}, abs/1109.2128.

\bibitem[{Frank and Wolfe(1956)}]{frank_wolfe_1956}
Marguerite Frank and Philip Wolfe. 1956.
\newblock \href {https://doi.org/10.1002/nav.3800030109} {An algorithm for
  quadratic programming}.
\newblock \emph{Naval Research Logistics Quarterly}, 3(1‐2):95--110.

\bibitem[{Gillick and Favre(2009)}]{gillick:2009}
Dan Gillick and Benoit Favre. 2009.
\newblock \href {https://www.aclweb.org/anthology/W09-1802} {A scalable global
  model for summarization}.
\newblock In \emph{Proceedings of the Workshop on Integer Linear Programming
  for Natural Language Processing}, pages 10--18, Boulder, Colorado.
  Association for Computational Linguistics.

\bibitem[{Goldstein et~al.(1999)Goldstein, Kantrowitz, Mittal, and
  Carbonell}]{Goldstein:1999}
Jade Goldstein, Mark Kantrowitz, Vibhu Mittal, and Jaime Carbonell. 1999.
\newblock \href {https://doi.org/10.1145/312624.312665} {Summarizing text
  documents: Sentence selection and evaluation metrics}.
\newblock In \emph{Proceedings of the 22Nd Annual International ACM SIGIR
  Conference on Research and Development in Information Retrieval}, SIGIR '99,
  pages 121--128, New York, NY, USA. ACM.

\bibitem[{Hahn and Mani(2000)}]{Hahn:2000}
Udo Hahn and Inderjeet Mani. 2000.
\newblock \href {https://doi.org/10.1109/2.881692} {The challenges of automatic
  summarization}.
\newblock \emph{Computer}, 33(11):29--36.

\bibitem[{Jaggi(2013)}]{jaggi2013revisiting}
Martin Jaggi. 2013.
\newblock Revisiting frank-wolfe: Projection-free sparse convex optimization.
\newblock In \emph{ICML (1)}, pages 427--435.

\bibitem[{Kedzie et~al.(2018)Kedzie, McKeown, and
  Daumé~III}]{kedzie_content_2018}
Chris Kedzie, Kathleen McKeown, and Hal Daumé~III. 2018.
\newblock \href {https://doi.org/10.18653/v1/D18-1208} {Content {Selection} in
  {Deep} {Learning} {Models} of {Summarization}}.
\newblock In \emph{Proceedings of the 2018 {Conference} on {Empirical}
  {Methods} in {Natural} {Language} {Processing}}, pages 1818--1828, Brussels,
  Belgium. Association for Computational Linguistics.

\bibitem[{Kupiec et~al.(1995)Kupiec, Pedersen, and Chen}]{Kupiec:1995}
Julian Kupiec, Jan Pedersen, and Francine Chen. 1995.
\newblock \href {https://doi.org/10.1145/215206.215333} {A trainable document
  summarizer}.
\newblock In \emph{Proceedings of the 18th Annual International ACM SIGIR
  Conference on Research and Development in Information Retrieval}, SIGIR '95,
  pages 68--73, New York, NY, USA. ACM.

\bibitem[{Lin(2004)}]{lin:2004}
Chin-Yew Lin. 2004.
\newblock Rouge: A package for automatic evaluation of summaries.
\newblock In \emph{Text Summarization Branches Out: Proceedings of the ACL-04
  Workshop}, pages 74--81, Barcelona, Spain. Association for Computational
  Linguistics.

\bibitem[{Luhn(1958)}]{luhn}
H.~P. Luhn. 1958.
\newblock \href {https://doi.org/10.1147/rd.22.0159} {The automatic creation of
  literature abstracts}.
\newblock \emph{IBM J. Res. Dev.}, 2(2):159--165.

\bibitem[{McDonald(2007)}]{McDonald:2007}
Ryan McDonald. 2007.
\newblock A study of global inference algorithms in multi-document
  summarization.
\newblock In \emph{Proceedings of the 29th European Conference on IR Research},
  ECIR'07, pages 557--564, Berlin, Heidelberg. Springer-Verlag.

\bibitem[{Mihalcea and Tarau(2004)}]{mihalcea-tarau-2004-textrank}
Rada Mihalcea and Paul Tarau. 2004.
\newblock \href {https://www.aclweb.org/anthology/W04-3252} {{T}ext{R}ank:
  Bringing order into text}.
\newblock In \emph{Proceedings of the 2004 Conference on Empirical Methods in
  Natural Language Processing}, pages 404--411, Barcelona, Spain. Association
  for Computational Linguistics.

\bibitem[{Nallapati et~al.(2016)Nallapati, Xiang, and Zhou}]{Nallapati:2016}
Ramesh Nallapati, Bing Xiang, and Bowen Zhou. 2016.
\newblock \href {http://arxiv.org/abs/1602.06023} {Sequence-to-sequence rnns
  for text summarization}.
\newblock \emph{CoRR}, abs/1602.06023.

\bibitem[{Nallapati et~al.(2017)Nallapati, Zhai, and
  Zhou}]{nallapati_summarunner_2017}
Ramesh Nallapati, Feifei Zhai, and Bowen Zhou. 2017.
\newblock {SummaRuNNer}: a recurrent neural network based sequence model for
  extractive summarization of documents.
\newblock In \emph{Proceedings of the {Thirty}-{First} {AAAI} {Conference} on
  {Artificial} {Intelligence}}, {AAAI}'17, pages 3075--3081, San Francisco,
  California, USA. AAAI Press.

\bibitem[{Ng and Abrecht(2015)}]{ng:2015}
Jun-Ping Ng and Viktoria Abrecht. 2015.
\newblock \href {https://doi.org/10.18653/v1/D15-1222} {Better summarization
  evaluation with word embeddings for {ROUGE}}.
\newblock In \emph{Proceedings of the 2015 Conference on Empirical Methods in
  Natural Language Processing}, pages 1925--1930, Lisbon, Portugal. Association
  for Computational Linguistics.

\bibitem[{Ouyang et~al.(2009)Ouyang, Li, Wei, and Lu}]{Ouyang:2009}
You Ouyang, Wenjie Li, Furu Wei, and Qin Lu. 2009.
\newblock Learning similarity functions in graph-based document summarization.
\newblock In \emph{Computer Processing of Oriental Languages. Language
  Technology for the Knowledge-based Economy}, pages 189--200, Berlin,
  Heidelberg. Springer Berlin Heidelberg.

\bibitem[{Paice(1990)}]{Paice:1990}
C.~D. Paice. 1990.
\newblock \href {https://doi.org/10.1016/0306-4573(90)90014-S} {Constructing
  literature abstracts by computer: Techniques and prospects}.
\newblock \emph{Inf. Process. Manage.}, 26(1):171--186.

\bibitem[{Peng et~al.(2019)Peng, Parikh, Faruqui, Dhingra, and Das}]{Peng:2019}
Hao Peng, Ankur~P. Parikh, Manaal Faruqui, Bhuwan Dhingra, and Dipanjan Das.
  2019.
\newblock \href {http://arxiv.org/abs/1904.04428} {Text generation with
  exemplar-based adaptive decoding}.
\newblock \emph{CoRR}, abs/1904.04428.

\bibitem[{Qian and Liu(2013)}]{liu:2013}
Xian Qian and Yang Liu. 2013.
\newblock \href {https://www.aclweb.org/anthology/D13-1156} {Fast joint
  compression and summarization via graph cuts}.
\newblock In \emph{Proceedings of the 2013 Conference on Empirical Methods in
  Natural Language Processing}, pages 1492--1502, Seattle, Washington, USA.
  Association for Computational Linguistics.

\bibitem[{Robertson and Zaragoza(2009)}]{Robertson:2009}
Stephen Robertson and Hugo Zaragoza. 2009.
\newblock \href {https://doi.org/10.1561/1500000019} {The probabilistic
  relevance framework: Bm25 and beyond}.
\newblock \emph{Found. Trends Inf. Retr.}, 3(4):333--389.

\bibitem[{Rush et~al.(2015)Rush, Chopra, and Weston}]{rush:2015}
Alexander~M. Rush, Sumit Chopra, and Jason Weston. 2015.
\newblock \href {https://doi.org/10.18653/v1/D15-1044} {A neural attention
  model for abstractive sentence summarization}.
\newblock In \emph{Proceedings of the 2015 Conference on Empirical Methods in
  Natural Language Processing}, pages 379--389, Lisbon, Portugal. Association
  for Computational Linguistics.

\bibitem[{Saggion and Poibeau(2012)}]{saggion:2012}
Horacio Saggion and Thierry Poibeau. 2012.
\newblock \href {https://hal.archives-ouvertes.fr/hal-00782442} {{Automatic
  Text Summarization: Past, Present and Future}}.
\newblock In R.~Yangarber T.~Poibeau; H. Saggion. J.~Piskorski, editor,
  \emph{{Multi-source, Multilingual Information Extraction and Summarization}},
  Theory and Applications of Natural Language Processing, pages 3--13.
  {Springer}.

\bibitem[{See et~al.(2017)See, Liu, and Manning}]{see:2017}
Abigail See, Peter~J. Liu, and Christopher~D. Manning. 2017.
\newblock \href {https://doi.org/10.18653/v1/P17-1099} {Get to the point:
  Summarization with pointer-generator networks}.
\newblock In \emph{Proceedings of the 55th Annual Meeting of the Association
  for Computational Linguistics (Volume 1: Long Papers)}, pages 1073--1083,
  Vancouver, Canada. Association for Computational Linguistics.

\bibitem[{Shao et~al.(2017)Shao, Gouws, Britz, Goldie, Strope, and
  Kurzweil}]{shao_generating_2017}
Louis Shao, Stephan Gouws, Denny Britz, Anna Goldie, Brian Strope, and Ray
  Kurzweil. 2017.
\newblock \href {http://arxiv.org/abs/1701.03185} {Generating {High}-{Quality}
  and {Informative} {Conversation} {Responses} with {Sequence}-to-{Sequence}
  {Models}}.
\newblock \emph{arXiv:1701.03185 [cs]}.
\newblock ArXiv: 1701.03185.

\bibitem[{Wang et~al.(2017)Wang, Zhao, Li, Ge, and
  Tang}]{wang_integrating_2017}
Shuai Wang, Xiang Zhao, Bo~Li, Bin Ge, and Daquan Tang. 2017.
\newblock \href {https://doi.org/10.1109/BigDataCongress.2017.46} {Integrating
  {Extractive} and {Abstractive} {Models} for {Long} {Text} {Summarization}}.
\newblock In \emph{2017 {IEEE} {International} {Congress} on {Big} {Data}
  ({BigData} {Congress})}, pages 305--312.

\bibitem[{Woodsend and Lapata(2011)}]{woodsend:2011}
Kristian Woodsend and Mirella Lapata. 2011.
\newblock \href {https://www.aclweb.org/anthology/D11-1038} {Learning to
  simplify sentences with quasi-synchronous grammar and integer programming}.
\newblock In \emph{Proceedings of the 2011 Conference on Empirical Methods in
  Natural Language Processing}, pages 409--420, Edinburgh, Scotland, UK.
  Association for Computational Linguistics.

\bibitem[{Xiao and Carenini(2019)}]{xiao_extractive_2019}
Wen Xiao and Giuseppe Carenini. 2019.
\newblock \href {http://arxiv.org/abs/1909.08089} {Extractive {Summarization}
  of {Long} {Documents} by {Combining} {Global} and {Local} {Context}}.
\newblock \emph{arXiv:1909.08089 [cs]}.
\newblock ArXiv: 1909.08089.

\bibitem[{Yao et~al.(2015)Yao, Wan, and Xiao}]{Yao:2015}
Jin-ge Yao, Xiaojun Wan, and Jianguo Xiao. 2015.
\newblock \href {http://dl.acm.org/citation.cfm?id=2832415.2832441}
  {Compressive document summarization via sparse optimization}.
\newblock In \emph{Proceedings of the 24th International Conference on
  Artificial Intelligence}, IJCAI'15, pages 1376--1382. AAAI Press.

\bibitem[{You et~al.(2016)You, Robinson, and Vidal}]{you_scalable_2016}
Chong You, Daniel~P. Robinson, and Rene Vidal. 2016.
\newblock \href {http://arxiv.org/abs/1507.01238} {Scalable {Sparse} {Subspace}
  {Clustering} by {Orthogonal} {Matching} {Pursuit}}.
\newblock \emph{arXiv:1507.01238 [cs, stat]}.
\newblock ArXiv: 1507.01238.

\bibitem[{Zeng et~al.(2016)Zeng, Luo, Fidler, and Urtasun}]{Zeng:2016}
Wenyuan Zeng, Wenjie Luo, Sanja Fidler, and Raquel Urtasun. 2016.
\newblock \href {http://arxiv.org/abs/1611.03382} {Efficient summarization with
  read-again and copy mechanism}.
\newblock \emph{CoRR}, abs/1611.03382.

\end{thebibliography}
